\documentclass[letterpaper,journal]{IEEEtran}
\usepackage{amsmath,amsfonts,amssymb,amsthm}
\usepackage{algorithm}
\usepackage{algpseudocode}
\algnewcommand\Input{\item[\textbf{Input:}]}
\algnewcommand\Output{\item[\textbf{Output:}]}
\usepackage{array}
\usepackage{textcomp}
\usepackage{stfloats}
\usepackage{url}
\usepackage{verbatim}
\usepackage{graphicx}
\usepackage{wrapfig}
\usepackage{subfigure}
\usepackage{multirow}
\usepackage{booktabs}
\usepackage{cite}
\usepackage{lipsum}
\usepackage{mathrsfs}
\usepackage{enumitem}
\usepackage{bm}
\usepackage{bbm}
\usepackage{soul}
\usepackage{nicefrac}
\usepackage{microtype}
\usepackage{mathtools}
\usepackage{color}
\usepackage{xcolor}
\usepackage{hyperref}
\usepackage[normalem]{ulem}


\usepackage{booktabs} 
\usepackage[utf8]{inputenc}     
\usepackage{amssymb} 

\begin{document}

\title{Domain-Aware Hierarchical Contrastive Learning for Semi-Supervised Generalization Fault Diagnosis}

\author{Junyu Ren, Wensheng Gan*, Philip S. Yu,~\IEEEmembership{Life Fellow,~IEEE}

\thanks{This research was supported in part by the National Natural Science Foundation of China (No. 62272196) and Guangzhou Basic and Applied Basic Research Foundation (No. 2024A04J9971).} 

\thanks{Junyu Ren and Wensheng Gan are with the College of Cyber Security, Jinan University, Guangzhou 510632, China. (E-mail: renjunyu193@gmail.com, wsgan001@gmail.com)} 
	
\thanks{Philip S. Yu is with the Department of Computer Science, University of Illinois Chicago, Chicago, USA. (E-mail: psyu@uic.edu)}

\thanks{Corresponding author: Wensheng Gan}
}

\maketitle

\begin{abstract}
Fault diagnosis under unseen operating conditions remains highly challenging when labeled data are scarce. Semi-supervised domain generalization fault diagnosis (SSDGFD) provides a practical solution by jointly exploiting labeled and unlabeled source domains. However, existing methods still suffer from two coupled limitations. First, pseudo-labels for unlabeled domains are typically generated primarily from knowledge learned on the labeled source domain, which neglects domain-specific geometric discrepancies and thus induces systematic cross-domain pseudo-label bias. Second, unlabeled samples are commonly handled with a hard accept-or-discard strategy, where rigid thresholding causes imbalanced sample utilization across domains, while hard-label assignment for uncertain samples can easily introduce additional noise. To address these issues, we propose a unified framework termed domain-aware hierarchical contrastive learning (DAHCL) for SSDGFD. Specifically, DAHCL introduces a domain-aware learning (DAL) module to explicitly capture source-domain geometric characteristics and calibrate pseudo-label predictions across heterogeneous source domains, thereby mitigating cross-domain bias in pseudo-label generation. In addition, DAHCL develops a hierarchical contrastive learning (HCL) module that combines dynamic confidence stratification with fuzzy contrastive supervision, enabling uncertain samples to contribute to representation learning without relying on unreliable hard labels. In this way, DAHCL jointly improves the quality of supervision and the utilization of unlabeled samples. Furthermore, to better reflect practical industrial scenarios, we incorporate engineering noise into the SSDGFD evaluation protocol. Extensive experiments on three benchmark datasets under severe noise and substantial domain shifts demonstrate that DAHCL consistently outperforms advanced SSDGFD baselines, exhibiting superior robustness and domain generalization capabilities. The code is publicly available at \url{https://github.com/JYREN-Source/DAHCL}.
\end{abstract}


\section{Introduction} \label{sec:Introduction}

Mechanical systems are fundamental to modern industry, including manufacturing, transportation, and wind energy \cite{fan2025novel, yi2025vibrmamba}. Faults in such systems may not only reduce production efficiency but also trigger serious safety incidents \cite{qi2025large}. Therefore, reliable fault diagnosis is essential for the safe and stable operation of industrial equipment. Driven by its powerful representation learning capability, deep learning (DL) has achieved remarkable success in intelligent fault diagnosis (IFD) \cite{lecun2015deep,chen2025review}. However, most DL-based methods are developed under the assumption that training and test data follow the same distribution. In practical industrial scenarios, variations in operating conditions often induce significant domain shifts, which can severely degrade diagnostic performance \cite{zhao2024domain}. To alleviate this issue, transfer learning has been widely introduced into fault diagnosis \cite{misbah2024fault,wang2025progressive}. Among existing paradigms, domain adaptation (DA)-based methods \cite{liu2025universal,ren2025global} improve cross-domain performance by leveraging target-domain data during training, but their dependence on target-domain accessibility limits their practical deployment. By contrast, domain generalization fault diagnosis (DGFD) \cite{zhao2024domain} learns only from source-domain data and aims to generalize to unseen target domains, making it more suitable for real-world applications. In particular, multi-source DGFD exploits multiple labeled source domains to learn domain-generalizable representations and has shown promising performance under varying operating conditions \cite{yang2025enhancing}. Nevertheless, its supervised learning paradigm still relies heavily on sufficient high-quality annotations, which are often expensive and difficult to obtain in industrial practice \cite{xiao2025domain}.

Semi-supervised domain generalization fault diagnosis (SSDGFD) provides a more practical alternative for fault diagnosis under unseen operating conditions when labeled data are scarce. It aims to learn a diagnosis model with cross-domain generalization capability by jointly exploiting a small amount of labeled source-domain data and abundant unlabeled source-domain data \cite{cui2026two}. Compared with supervised DGFD, the core challenge of SSDGFD lies in constructing sufficiently reliable supervisory signals for unlabeled source domains in the presence of both domain shifts and incomplete annotations \cite{jiang2025semi}. To this end, existing studies typically combine domain-level alignment with category-level regularization, such as pseudo-labeling, to provide approximate supervision for unlabeled samples \cite{liao2020deep,ren2023domain}. Recent efforts have further improved this paradigm from different perspectives, including domain-invariant feature extraction \cite{zhao2023mutual}, feature disentanglement \cite{song2024contrast}, class-prior modeling \cite{ying2025semi}, pseudo-label filtering \cite{wei2025domain}, and augmentation-based consistency enhancement \cite{cui2026two}. Although these methods have achieved encouraging progress, they still predominantly rely on the conventional paradigm of domain-invariant representation learning, with category-level constraints serving mainly as auxiliary supervision for unlabeled samples. 

Nevertheless, existing SSDGFD methods still face three tightly coupled challenges. 
(1) \textit{Cross-domain bias in pseudo-label generation}: most methods generate pseudo-labels mainly according to the decision boundary learned from the labeled source domain, while overlooking the geometric characteristics of each unlabeled source domain. As a result, the generated pseudo-labels are implicitly biased toward the labeled domain and can become systematically unreliable when the unlabeled domains exhibit varying degrees of similarity to it. This issue is particularly severe under large domain shifts, where noisy pseudo-labels may be repeatedly reinforced during training. 
(2) \textit{Binary sample selection leads to inefficient use of unlabeled data}: existing category-level supervision strategies for unlabeled samples often rely on a hard confidence criterion to determine whether a sample should be trusted or ignored. However, a fixed threshold is inherently insensitive to heterogeneous transferability across source domains, often causing samples from domains closer to the labeled source to dominate training while those from more distant domains remain underutilized. More importantly, uncertain samples near the confidence threshold are not truly useless: forcing them into hard labels introduces noise, whereas discarding them entirely wastes potentially informative semantic structure. 
(3) \textit{Insufficient evaluation under realistic noisy conditions}: most existing SSDGFD studies are conducted under relatively clean benchmark settings. However, vibration signals acquired in real industrial environments are often subject to substantial engineering noise \cite{chen2025noise}. Such noise not only blurs class boundaries but also further magnifies pseudo-label bias and the imbalance in unlabeled sample utilization across domains.

\begin{figure}[htbp]
    \centering
    \includegraphics[width=1\linewidth]{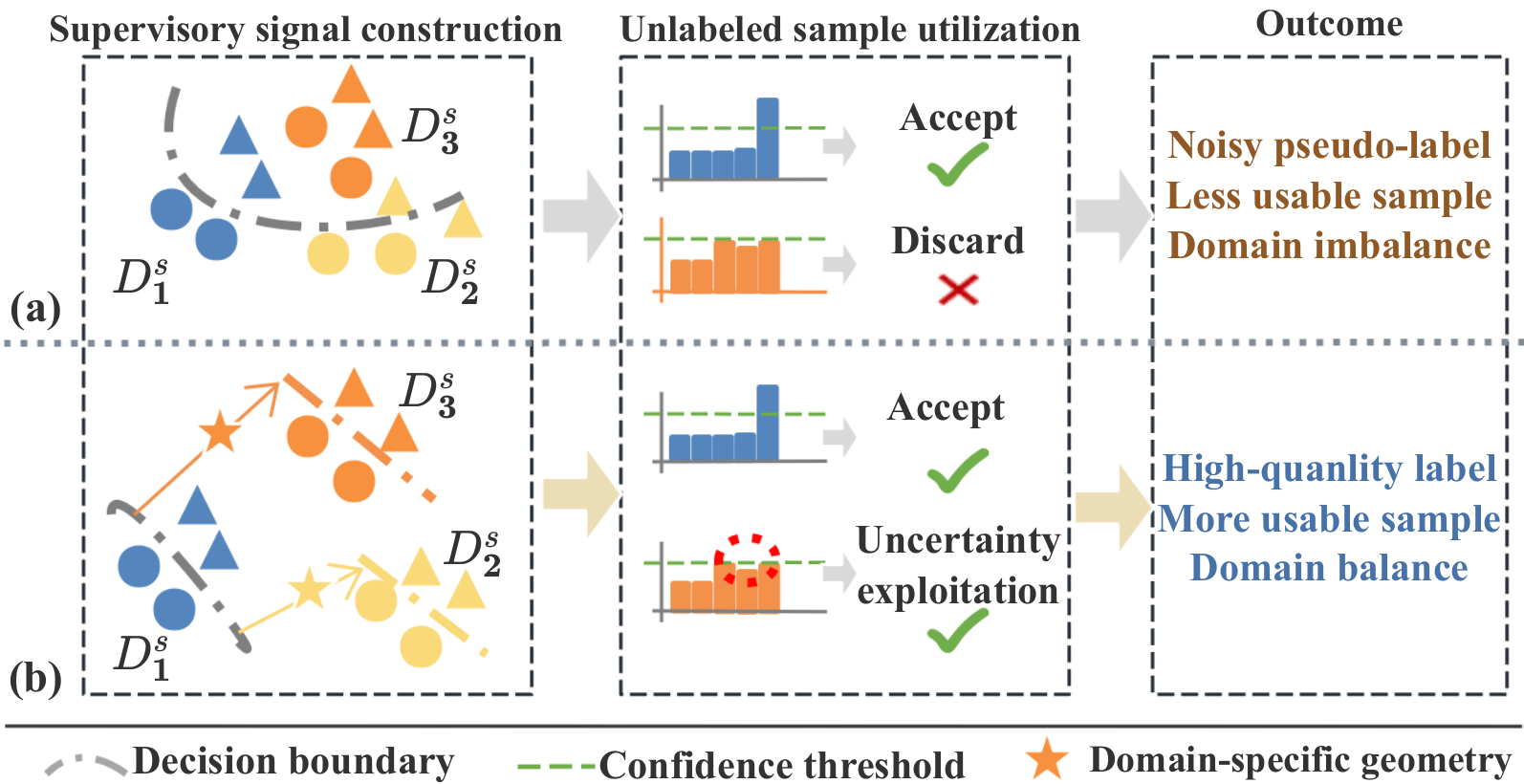}
    \caption{Visual illustration of differences between (a) previous works and (b) our method.}
    \label{fig.innovation}
\end{figure}

As illustrated in Fig.~\ref{fig.innovation}, the above limitations stem from a common issue: existing SSDGFD methods mainly emphasize domain-invariant learning, but underexploit two important sources of information contained in unlabeled source domains, namely domain-specific geometric characteristics and uncertainty-aware semantic cues. In other words, domain-specific characteristics are often treated merely as nuisance variations to be suppressed, even though they can provide valuable evidence for correcting cross-domain pseudo-label bias. Meanwhile, uncertain samples are usually handled in a hard accept-or-discard manner, although their prediction distributions still contain useful weak semantic information.

To address these issues, we propose \textit{domain-aware hierarchical contrastive learning} (DAHCL), a unified framework that improves SSDGFD from two complementary perspectives. First, DAHCL introduces a \textit{domain-aware learning} scheme to explicitly encode source-domain geometric characteristics into the prediction process, thereby reducing the systematic bias in pseudo-labels across heterogeneous source domains. Second, DAHCL develops \textit{hierarchical contrastive learning}, which replaces rigid binary selection with stratified supervision and fuzzy contrastive learning, enabling uncertain samples to contribute to representation learning without relying on unreliable hard labels. In this way, DAHCL jointly achieves bias correction and uncertainty-aware utilization of unlabeled data, leading to more robust generalization under severe noise and large domain shifts. The main contributions of this paper are summarized as follows:

\begin{itemize}
    \item We propose a unified DAHCL framework for SSDGFD that rethinks the role of unlabeled source domains from two overlooked aspects: domain-specific geometry and uncertainty-aware supervision. Unlike existing methods that mainly pursue domain invariance and hard pseudo-label filtering, DAHCL jointly addresses cross-domain pseudo-label bias and insufficient utilization of uncertain unlabeled samples.

    \item We design a \textit{domain-aware learning} (DAL) module to explicitly model source-domain geometric characteristics and build domain-aware experts for pseudo-label calibration. By constraining predictions to be consistent with domain-specific distribution patterns, DAL alleviates the systematic bias caused by transferring labeled-domain decision knowledge directly to heterogeneous unlabeled domains, and further supports adaptive expert selection for unseen target domains during inference.

    \item We develop \textit{hierarchical contrastive learning} (HCL) to replace one-shot hard sample filtering with dynamic confidence stratification and fuzzy contrastive supervision. This design allows uncertain samples to provide useful supervisory signals in a soft manner, improving utilization of unlabeled data and mitigating cross-domain imbalance without introducing excessive hard-label noise.

    \item We extend SSDGFD to a more realistic noisy evaluation setting by incorporating engineering noise into both training and testing. This protocol better reflects practical fault diagnosis scenarios and provides a more rigorous benchmark for assessing robustness under severe signal contamination.

    \item Extensive experiments on three benchmark datasets demonstrate that DAHCL consistently outperforms existing advanced SSDGFD methods under severe noise and large domain shifts, showing stronger robustness and generalization capability.
\end{itemize}

This paper is organized as follows: Section \ref{sec:Related Works} reviews related work. Section \ref{sec:algorithm} presents the proposed method, and Section \ref{sec: experiments} shows experimental results and analysis. Finally, Section \ref{sec:conclusion} provides conclusions and future research directions.

\section{Related Work}  \label{sec:Related Works}

\subsection{Multi-source DG Fault Diagnosis}

Multi-source domain generalization fault diagnosis (MSDGFD) aims to train a diagnostic model on multiple known source domains and generalize the learned knowledge to unseen operating conditions without access to target-domain data. Owing to its practical significance, this problem has attracted increasing attention in recent years. The key objective is to learn feature representations from multiple labeled source domains that are both class-discriminative and stable across domains \cite{huang2025uddgn,yang2025enhancing}. Existing MSDGFD methods can be broadly categorized into two lines of research. The first line improves generalization by expanding the training distribution at the data level, such as through data augmentation, domain augmentation, or cross-domain interpolation. For example, Shi \emph{et al.} \cite{shi2023domain} proposed a domain augmentation generalization network that enlarges the support of the source-domain distribution by constructing augmented domains and incorporating adversarial training. Guan \emph{et al.} \cite{guan2025domain} further generated linear intermediate domains between different source domains to enhance the continuity of the training distribution and improve model adaptability to out-of-distribution samples. The second line focuses on domain-invariant representation learning, which aims to extract transferable shared diagnostic knowledge from multiple source domains via adversarial alignment, distribution discrepancy minimization, or metric learning. Specifically, Gao \emph{et al.} \cite{gao2025multiple} characterized multi-source domain discrepancies by exploiting the prediction inconsistency among multiple domain-specific classifiers and achieved finer-grained global feature alignment with the Wasserstein distance. In another work, Gao \emph{et al.} \cite{gao2024domain} introduced feature disentanglement to explicitly separate fault-related representations from domain-related components, thereby obtaining purer and more stable domain-invariant features. Moreover, causal representation learning has also been introduced into MSDGFD, where features are decomposed into causal and non-causal factors to reduce the adverse effect of spurious correlations on generalization performance \cite{ma2024causality}. Beyond feature learning, some studies have attempted to improve multi-source generalization from the perspective of training paradigms and model organization, such as meta-learning frameworks based on gradient alignment and semantic matching \cite{ren2023meta}, and ensemble learning methods with collaborative multi-branch modeling \cite{xiao2025domain2}.

Although MSDGFD methods have significantly improved fault diagnosis performance under unseen operating conditions, they are generally built upon the assumption of sufficient labeled data, which does not conform to realistic industrial scenarios. Therefore, effectively exploiting unlabeled source-domain data has become a key issue in advancing DGFD toward practical applications.

\subsection{Semi-supervised DG Fault Diagnosis}

Semi-supervised domain generalization fault diagnosis (SSDGFD) assumes that only a portion of the source-domain samples are labeled. Its core challenge lies in how to provide reliable category-level supervision for unlabeled source domains while ensuring effective sample utilization. Since it enables the exploitation of massive yet difficult-to-annotate industrial data, SSDGFD is of greater practical value in label-scarce scenarios. Most existing SSDGFD methods follow a general framework that combines global domain alignment with semi-supervised learning. For instance, Liao \emph{et al.} \cite{liao2020deep} integrated a Wasserstein generative adversarial network with a pseudo-labeling strategy for training. Li \emph{et al.} \cite{li2022new} used the prediction discrepancy between two classifiers to characterize cross-domain representation similarity. Ren \emph{et al.} \cite{ren2023domain} achieved global domain alignment through adversarial training and further imposed category-level constraints by incorporating a pseudo-label-based center loss. Subsequently, Ren \emph{et al.} \cite{ren2023domain2} proposed DIFFN, which extracts discriminative features and domain-invariant features through two separate branches and then performs joint fusion modeling. These methods mainly focus on mitigating domain shift at the global distribution level, while their characterization of category structure is relatively coarse or even absent. Building upon this line of research, subsequent studies began to incorporate feature disentanglement, contrastive learning, and sample selection mechanisms. Zhao \emph{et al.} \cite{zhao2023mutual} proposed a multi-assistance semi-supervised DG network, in which domain alignment is mainly used for pseudo-label assignment, while low-rank decomposition is exploited to guide the training of domain-specific and domain-invariant classifiers. Song \emph{et al.} \cite{song2024contrast} proposed a domain-specificity removal network, which obtains purer domain-invariant representations by removing domain-private features and further enhances category-level constraints via proxy-based contrastive learning. Compared with earlier methods, these approaches further improve the extraction of domain-invariant features. However, they fail to account for the similarity differences between unlabeled and labeled source domains, which may easily lead to noisy pseudo labels and inter-domain imbalance in the number of usable samples. To address the above issues, recent studies have further improved SSDGFD from the perspectives of sample utilization and category constraint design. Wei \emph{et al.} \cite{wei2025domain} quantified the degree of imbalance according to the number of pseudo-labeled samples in each domain and dynamically adjusted the confidence threshold to improve sample utilization. However, this strategy essentially increases the number of usable samples by relaxing the threshold and therefore still suffers from the risk of introducing more noise. Jiang \emph{et al.} \cite{jiang2025semi} injected category constraints into the network weight generation process, thereby avoiding explicit reliance on pseudo labels. Cui \emph{et al.} \cite{cui2026two} proposed a two-stage semi-supervised domain generalization network, which first improves pseudo-label quality through MMD-driven domain adaptation and then enhances generalization capability using Mixup. Nevertheless, these methods still rely solely on knowledge from labeled source domains to impose category constraints.

Overall, although existing SSDGFD methods have achieved encouraging progress, they still suffer from three main limitations. First, the modeling of category structure in unlabeled source domains remains insufficient. Pseudo-label generation or implicit category constraints still primarily depend on knowledge from labeled source domains, without explicitly characterizing the geometric structure of unlabeled domains or the similarity differences across domains, which may easily induce cross-domain bias. Second, unlabeled sample utilization still largely follows fixed-threshold and hard-selection mechanisms, which essentially amount to a binary ``accept-or-discard'' strategy, making it difficult to balance the number of usable samples across different source domains and to properly exploit uncertain samples. Third, most existing studies are conducted under relatively ideal data conditions and have not adequately considered the influence of engineering noise in rotating machinery systems on SSDGFD performance.

\section{Proposed Method} \label{sec:algorithm}
\subsection{Problem Definition} \label{sec:Problem Definition}

In SSDGFD, data are collected from multiple operating conditions, resulting in multiple source domains with distribution discrepancies. Let $\{\mathcal{D}_m^s\}_{m=1}^{M}$ denote the set of $M$ available source domains, where only one source domain $\mathcal{D}_1^s=\{(\mathbf{x}_{1,i}^s,y_{1,i})\}_{i=1}^{n_1^s}$ is fully labeled, while the remaining source domains $\mathcal{D}_m^s=\{\mathbf{x}_{m,i}^s\}_{i=1}^{n_m^s},\, m=2,\dots,M$ are unlabeled. Here, $\mathbf{x}_{m,i}^s$ denotes the $i$-th sample from the $m$-th source domain and $y_{1,i}\in\{1,\dots,K\}$ represents the corresponding fault category. During training, the target domain $\mathcal{D}^t=\{\mathbf{x}_i^t\}_{i=1}^{n^t}$ is completely inaccessible and is only used for performance evaluation in the testing phase. Although all domains share the same fault label space, their data distributions differ due to varying operating conditions, i.e., $P_1(\mathcal{X}_1^s)\neq \cdots \neq P_M(\mathcal{X}_M^s)\neq P_t(\mathcal{X}^t)$. The goal is to enable reliable fault identification on an unseen target domain using partially labeled source-domain data.

\subsection{Overview of DAHCL}

\begin{figure*}[htbp]
    \centering
    \includegraphics[width=1\linewidth]{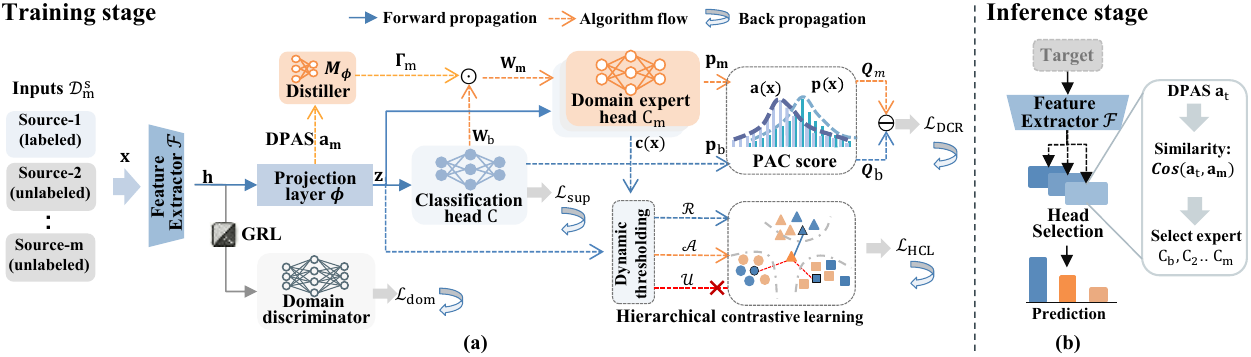}
    \caption{Overview of the proposed DAHCL.}
    \label{fig.overview}
\end{figure*}

The proposed DAHCL consists of a shared feature backbone and two key components: 1) \emph{domain-aware learning} (DAL), which exploits source-domain geometric characteristics to improve pseudo-label quality during training and enables adaptive expert selection at test time; and 2) \emph{hierarchical contrastive learning} (HCL), which stratifies unlabeled samples by confidence and imposes differentiated contrastive supervision to improve utilization of unlabeled data and alleviate cross-domain imbalance. As shown in Fig. \ref{fig.overview}, the overall framework contains four modules:
1) a feature extractor \(\mathcal{F}: \mathbf{x} \mapsto \mathbf{h} \in \mathbb{R}^{D_0}\), which extracts high-level representations from raw vibration signals;
2) a classifier \(G_C\), including a projection layer \(\phi: \mathbf{h} \mapsto \mathbf{z} \in \mathbb{R}^{d}\) and a linear classification head \(C: \mathbf{z} \mapsto \boldsymbol{\ell}_{\mathrm{base}} \in \mathbb{R}^{K}\), where \(\mathbf{z}\) denotes the feature embedding and \(\boldsymbol{\ell}_{\mathrm{base}}\) denotes the logits;
3) a domain-adversarial alignment module composed of a domain discriminator \(G_D: \mathbf{h} \mapsto \boldsymbol{\ell}_\mathrm{D} \in \mathbb{R}^{M}\) and a gradient reversal layer (GRL) \cite{ganin2016domain}, which performs global domain alignment; and
4) a domain knowledge distillation module \(M_{\phi}: \mathbb{R}^{K} \rightarrow \mathbb{R}^{K \times d}\), which generates domain-aware modulation matrices for expert construction. The detailed design is described below.

\subsection{Supervised Classification and Domain-Adversarial Alignment}

The feature extractor \(\mathcal{F}\) and the classifier \(G_C\) form the basic diagnosis backbone. For the labeled source domain \(\mathcal{D}_1^s\), the supervised classification loss is defined as
\begin{equation}
\mathcal{L}_{\mathrm{sup}}
=
\mathbb{E}_{(\mathbf{x},y)\sim \mathcal{D}_1^s}
\left[
\mathrm{CE}\bigl(C(\phi(\mathcal{F}(\mathbf{x}))),y\bigr)
\right],
\label{eq:CLS}
\end{equation}
where \(\mathrm{CE}(\cdot,\cdot)\) denotes the cross-entropy loss.

To preliminarily reduce domain shift, we further introduce a domain-adversarial alignment module consisting of \(G_D\) and a GRL. The domain discrimination loss over all source domains is given by
\begin{equation}
\mathcal{L}_{\mathrm{dom}}
=
\frac{1}{M}\sum_{m=1}^{M}
\mathbb{E}_{\mathbf{x}\sim \mathcal{D}_m^s}
\left[
\mathrm{CE}\bigl(G_D(\mathcal{F}(\mathbf{x})),m\bigr)
\right],
\label{eq:DOM}
\end{equation}
where \(m\) is the domain label. Through the GRL, \(G_D\) is optimized to minimize \(\mathcal{L}_{\mathrm{dom}}\), whereas \(\mathcal{F}\) receives a reversed gradient and is thus encouraged to learn coarse domain-invariant features \cite{ganin2016domain}. We further develop DAL and HCL on top of the above backbone.

\subsection{Domain-Aware Learning}
\label{sec:DAL}

In SSDGFD, pseudo-labels for unlabeled source domains are often generated solely based on the decision boundary learned from the labeled source domain. Under a substantial domain shift, such a strategy easily introduces systematic cross-domain bias and repeatedly reinforces it through noisy pseudo-labels. To address this issue, we propose \emph{domain-aware learning} (DAL), which explicitly embeds domain-specific geometric characteristics into the prediction process for bias correction. 

\subsubsection{Domain-specific geometry embedding}

We first define the \emph{domain-proxy affinity spectrum} (DPAS) to characterize each unlabeled source domain from a geometric perspective. For the base classifier, let
\(\mathbf{W}_{\mathrm{base}}=[\mathbf{w}_1,\ldots,\mathbf{w}_K]^{\top}\in\mathbb{R}^{K\times d}\),
where \(\mathbf{w}_k\) is the proxy of the \(k\)th fault category. For the \(m\)th unlabeled source domain \(\mathcal{D}_m^s\) (\(m\in\{2,\ldots,M\}\)), the instantaneous DPAS on a mini-batch \(\mathcal{B}_m\) in training process is defined as
\begin{equation}
\mathbf{s}_m
=
\frac{1}{|\mathcal{B}_m|}
\sum_{\mathbf{x}\in\mathcal{B}_m}
\bigl[
\cos(\mathbf{z}(\mathbf{x}),\mathbf{w}_1),\ldots,
\cos(\mathbf{z}(\mathbf{x}),\mathbf{w}_K)
\bigr]^{\top},
\label{eq:dpas}
\end{equation}
To obtain a stable domain descriptor, we then update the domain descriptor by an exponential moving average method:
\begin{equation}
\mathbf{a}_m \leftarrow (1-\mu)\mathbf{a}_m+\mu\mathbf{s}_m,
\label{eq:dpas2}
\end{equation}
where \(\mu\) is the momentum coefficient. The resulting \(\mathbf{a}_m\in\mathbb{R}^{K}\) serves as a compact semantic descriptor of the \(m\)th unlabeled source domain, whose $k$th component represents the average cosine similarity between samples in the unlabeled source domain $\mathcal{D}_m^s$ and the proxy of class $k$. This geometry-aware metric, constructed based on affinity distributions, characterizes the feature tendency of a domain and reflects its distinctive distributional properties in the overall class space.

\subsubsection{Domain-aware expert classifier}

Although \(\mathbf{a}_m\) captures the overall geometric tendency of a source domain, a simple linear mapping is insufficient to encode such information into the classifier. Therefore, we introduce a lightweight domain knowledge distillation module \(M_{\phi}\) to adaptively encode the domain knowledge $\mathbf{a}_m$. Its output $\boldsymbol{\Gamma}_m^{\mathrm{raw}}$ is mapped to a modulation matrix via a residual connection and an activation function, which can be formulated as:

\begin{equation}
\boldsymbol{\Gamma}_{m}^{\mathrm{raw}}=M_{\phi}(\mathbf{a}_m), \qquad
\boldsymbol{\Gamma}_{m}
=
\mathbf{1}
+
\varepsilon \cdot \tanh\!\left(\boldsymbol{\Gamma}_{m}^{\mathrm{raw}}\right),
\end{equation}
where \(\varepsilon\) controls the modulation magnitude and \(\mathbf{1}\) is an all-one matrix. Based on this design, \(M-1\) domain-aware expert heads are constructed for the \(M-1\) unlabeled source domains. Specifically, the weight matrix of the \(m\)th expert is defined as
\begin{equation}
\mathbf{W}_m=\mathbf{W}_{\mathrm{base}}\odot \boldsymbol{\Gamma}_{m},
\end{equation}
where \(\odot\) denotes element-wise multiplication. The logits produced by the base classifier and the \(m\)th domain-aware expert are denoted by \(\boldsymbol{\ell}_{\mathrm{base}}(\mathbf{x})\) and \(\boldsymbol{\ell}_{m}(\mathbf{x})\), respectively.

\subsubsection{Domain-aware coherence regularization}
During backpropagation, we construct a \emph{domain-aware coherence regularization (DCR)} term that exploits domain geometric characteristics to constrain the predictions of domain experts. Specifically, for an unlabeled sample $\mathbf{x}$ with prediction probability vector $\mathbf{p}(\mathbf{x})$
and affinity vector $\mathbf{a}_m(\mathbf{x})$,
we define the prediction-affinity coherence score as:
\begin{equation}
Q\bigl(\mathbf{p}(\mathbf{x}),\mathbf{a}_m(\mathbf{x})\bigr)
=\mathbf{p}(\mathbf{x})^{\top}\mathbf{a}_m(\mathbf{x})
=\sum_{k=1}^{K} p_k(\mathbf{x})\,a_{m,k}(\mathbf{x}).
\end{equation}
This score measures the consistency between the classifier prediction distribution and the DPAS vector. Specifically, let $\mathbf{p}_{\mathrm{base}}(\mathbf{x}) = \mathrm{softmax}(\boldsymbol{\ell}_{\mathrm{base}}(\mathbf{x}))$
and $\mathbf{p}_{m}(\mathbf{x}) = \mathrm{softmax}(\boldsymbol{\ell}_{m}(\mathbf{x}))$
denote the prediction distributions of the base classifier and the current domain expert, respectively. Their corresponding coherence scores are
$Q_{\mathrm{base}}(\mathbf{x}) = Q(\mathbf{p}_{\mathrm{base}}(\mathbf{x}), \mathbf{a}_m(\mathbf{x}))$
and
$Q_{m}(\mathbf{x}) = Q(\mathbf{p}_{m}(\mathbf{x}), \mathbf{a}_m(\mathbf{x}))$,
respectively. Here, $Q_{\mathrm{base}}$ is treated as a fixed reference (with gradient detached). The optimization target of DCR is defined as the coherence violation cost:
\begin{equation}
  \mathcal{L}_{\text{DCR}}
  = \frac{1}{|\mathcal{B}_m|} \sum_{\mathbf{x} \in \mathcal{B}_m}
    \max\big(0,\, Q_{\mathrm{base}}(\mathbf{x}) - Q_{m}(\mathbf{x})\big),
      \label{eq:DCR}
\end{equation}
This loss drives the prediction distribution of the domain expert to be more coherent with the domain geometric vector than that of the base classifier, thereby allowing the model to adaptively adjust its decision boundary according to the geometric characteristics of the corresponding source domain and suppress semantic mismatch caused by domain shift.

\subsubsection{Test-time domain expert selection}
During inference, the domain expert classifier that best matches the target domain is selected by computing DPAS similarity scores, thereby adaptively improving diagnostic accuracy. Specifically, the DPAS representation of the target domain, denoted by $\mathbf{a}_t$, is updated online. For each incoming mini-batch $\mathcal{B}_t$, the instantaneous DPAS $\mathbf{s}_t$ is computed in the same manner as in eq. \eqref{eq:dpas},
and $\mathbf{a}_t$ is updated via exponential moving average as $\mathbf{a}_t \leftarrow (1-\mu)\,\mathbf{a}_t + \mu\,\mathbf{s}_t$. Subsequently, the DPAS vectors are $L_2$-normalized, and the similarity between the target domain and each source domain is measured by cosine similarity:
\begin{equation}
  \rho_m = \frac{\mathbf{a}_t^\top \mathbf{a}_m}{\|\mathbf{a}_t\|_2 \|\mathbf{a}_m\|_2}, \quad m \in \{2,\ldots,M\},
  \label{eq:expert_selection}
\end{equation}
According to the nearest-neighbor principle, the index of the source domain with the highest similarity is selected as the expert assignment for the target domain: $ m^* = \arg\max_{m} \rho_m$, and the final prediction is given by the corresponding domain expert classifier: $\hat{y} = \arg\max_k [\boldsymbol{\ell}_{m^*}(\mathbf{x})]_k$.

\subsection{Hierarchical contrastive learning}

Existing pseudo-label-based SSDGFD methods usually employ a single threshold to divide unlabeled samples into accepted and discarded sets. However, such a binary strategy cannot accommodate heterogeneous domain gaps, causing over-utilization of samples from domains close to the labeled source and under-utilization of those from distant domains, while also neglecting the useful semantic information carried by uncertain samples. To address this issue, we propose hierarchical contrastive learning (HCL), which combines dynamic cross-domain thresholding with stratified contrastive constraints to enable fine-grained confidence partitioning and differentiated supervision for unlabeled samples.

\subsubsection{Dynamic cross-domain thresholding}

At each iteration, unlabeled samples from all \(M-1\) unlabeled source domains are mixed into a batch \(\mathcal{B}_u\). For a sample \(\mathbf{x}\in\mathcal{D}_m^s\), its confidence score is computed from the corresponding domain-aware expert: $c(\mathbf{x})=\max_{k} p_{m,k}(\mathbf{x}),$
where \(p_{m,k}(\mathbf{x})\) is the \(k\)th entry of \(\mathbf{p}_m(\mathbf{x})\). Collect all confidence scores in \(\mathcal{B}_u\) as \(\{c_i\}_{i=1}^{|\mathcal{B}_u|}\), and sort them in ascending order as \(\{c_{(i)}\}_{i=1}^{|\mathcal{B}_u|}\), where \(c_{(1)}\leq \cdots \leq c_{(|\mathcal{B}_u|)}\).Then, two quantiles are computed as $q_{\mathrm{low}}=c_{(\lceil \eta_1|\mathcal{B}_u| \rceil)}$, $q_{\mathrm{mid}}=c_{(\lceil \eta_2|\mathcal{B}_u| \rceil)},$ where \(0<\eta_1<\eta_2<1\). If only quantiles are used, the thresholds will increase synchronously with the overall confidence level during late training, forcing a fixed proportion of samples to remain in low-confidence regions even when most samples have become reliable. To avoid this issue, we introduce two upper-bound thresholds \(\tau_1<\tau_2\), and define the actual thresholds as
$t_{\mathrm{low}}=\min(q_{\mathrm{low}},\tau_1)$, $t_{\mathrm{mid}}=\min(q_{\mathrm{mid}},\tau_2).$
Accordingly, unlabeled samples are partitioned into three subsets: $\mathcal{R} = \{\mathbf{x} \mid c(\mathbf{x}) \geq t_{\mathrm{mid}}\}$, $\mathcal{A} = \{\mathbf{x} \mid t_{\mathrm{low}} \leq c(\mathbf{x}) < t_{\mathrm{mid}}\}$, $\mathcal{U} = \{\mathbf{x} \mid c(\mathbf{x}) < t_{\mathrm{low}}\}$, where \(\mathcal{R}\), \(\mathcal{A}\), and \(\mathcal{U}\) denote the reliable, ambiguous, and unreliable partitions, respectively.

\begin{figure}[htbp]
    \centering
    \includegraphics[width=1\linewidth]{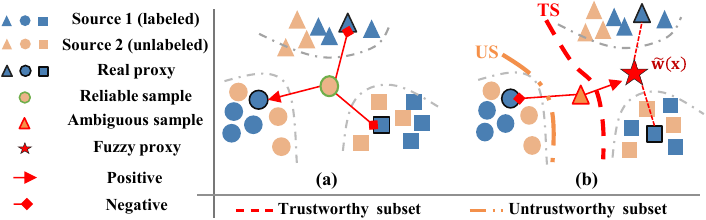}
    \caption{Illustration of the proposed hierarchical contrastive learning.}
    \label{fig.hcl}
\end{figure}

\subsubsection{Stratified contrastive constraints}
Based on the three-level partition, HCL imposes differentiated supervision on different subsets. For reliable samples \(\mathcal{R}\), standard proxy-based contrastive learning is performed. For unreliable samples \(\mathcal{U}\), no pseudo-label supervision or contrastive constraint is imposed to avoid noise propagation. Specifically, for uncertain samples \(\mathcal{A}\), we further design a fuzzy contrastive mechanism to exploit their soft semantic information. The feature embedding $\mathbf{z}$ is used to construct the contrastive metric space of HCL. To formulate the contrastive learning objective, the feature embedding $\mathbf{z}$ and the proxies $\mathbf{w}_k$ are first $L_2$-normalized:
$
\bar{\mathbf{z}}= \frac{\mathbf{z}}{\|\mathbf{z}\|_2}, \quad
\bar{\mathbf{w}}_k = \frac{\mathbf{w}_k}{\|\mathbf{w}_k\|_2}.
$
A temperature coefficient $T > 0$ is then introduced to define the unified temperature-scaled cosine similarity:
$s\!\left(\mathbf{z}, \mathbf{w}_k\right) = \frac{\bar{\mathbf{z}}^\top \bar{\mathbf{w}}_k}{T}.
$

For a reliable sample \(\mathbf{x}\in\mathcal{R}\), the pseudo-label is obtained according to the prediction distribution $\mathbf{p}_{m}(\mathbf{x})$ of the Domain-expert:
$\hat{y}(\mathbf{x}) = \arg\max_k p_{m,k}(\mathbf{x})$.
As illustrated in Fig. \ref{fig.hcl} (a), the positive sample of $\mathbf{x}$ is the proxy $\mathbf{w}_{\hat{y}(\mathbf{x})}$ corresponding to its predicted category, while the remaining $K-1$ class proxies are treated as negative samples. We employ the proxy-based InfoNCE \cite{oord2018representation} loss:
\begin{equation}
\mathcal{L}_{\mathcal{R}}(\mathbf{x})
= -\log
\frac{\exp\!\bigl(s(\mathbf{z}, \mathbf{w}_{\hat{y}(\mathbf{x})})\bigr)}
{\sum_{k=1}^{K} \exp\!\bigl(s(\mathbf{z}, \mathbf{w}_k)\bigr)}.
\end{equation}

For an uncertain sample \(\mathbf{x}\in\mathcal{A}\), its one-hot pseudo-labels are unreliable, and premature hard assignment may erase potentially informative secondary category information; meanwhile, completely discarding such samples would hinder the learning of critical features. Inspired by fuzzy theory \cite{zadeh1965fuzzy}, we divide the predicted categories into Trustworthy Classes and Untrustworthy Classes, and construct a fuzzy proxy from the trustworthy class proxies as the sole positive sample, while the remaining untrustworthy class proxies are treated as negative samples, as illustrated in Fig. \ref{fig.hcl} (b).  
Specifically, the top-3 classes $\mathcal{T}_3(\mathbf{x})$ whose predicted probabilities are higher than the random-guessing prior $1/K$ are defined as the trustworthy classes, i.e., the trustworthy class subset is
$\mathcal{C}(\mathbf{x})
= \bigl\{ k \in \mathcal{T}_3(\mathbf{x}) \mid p_{m,k}(\mathbf{x}) > 1/K \bigr\}$.
Probability normalization is then performed over $\mathcal{C}(\mathbf{x})$ to obtain the weights:
\begin{equation}
\omega_k(\mathbf{x})
=\frac{p_{m,k}(\mathbf{x})}
{\sum_{j\in\mathcal{C}(\mathbf{x})}p_{m,j}(\mathbf{x})},
\qquad  k\in\mathcal{C}(\mathbf{x}).
\end{equation}
The fuzzy proxy is obtained by probability-weighted aggregation over the trustworthy class proxies:
\begin{equation}
  \tilde{\mathbf{w}}(\mathbf{x})
  = \sum_{k \in \mathcal{C}(\mathbf{x})} \omega_k(\mathbf{x})\, \mathbf{w}_k.
  \label{eq:fuzzy_proxy}
\end{equation}
The untrustworthy class subset is defined as the complement of the trustworthy class subset:
$\mathcal{N}(\mathbf{x}) = \{1, \dots, K\} \setminus \mathcal{C}(\mathbf{x})$.
The denominator performs softmax normalization over the positive sample and all negative samples. Accordingly, the contrastive loss for uncertain samples is defined as:
\begin{equation}
\mathcal{L}_{\mathcal{A}}(\mathbf{x})
= -\log
\frac{\exp\!\bigl(s(\mathbf{z}, \tilde{\mathbf{w}}(\mathbf{x}))\bigr)}
{\exp\!\bigl(s(\mathbf{z}, \tilde{\mathbf{w}}(\mathbf{x}))\bigr) + \sum\limits_{k \in \mathcal{N}(\mathbf{x})}
\exp\!\bigl(s(\mathbf{z}, \mathbf{w}_k)\bigr)}.
\end{equation}
This design prevents uncertain samples from being forcibly clustered into one-hot categories. Instead, they are first pulled toward the weighted center of the trustworthy class proxies. As the model gradually converges, their prediction distributions naturally become sharper, thereby enabling a smooth transition from fuzzy supervision to deterministic supervision.

In summary, the overall loss of HCL can be expressed as:
\begin{equation}
\mathcal{L}_{\mathrm{HCL}} = \frac{1}{|\mathcal{R}|}\sum_{\mathbf{x} \in \mathcal{R}} \mathcal{L}_{\mathcal{R}}(\mathbf{x})
+ \frac{1}{|\mathcal{A}|}\sum_{\mathbf{x} \in \mathcal{A}} \mathcal{L}_{\mathcal{A}}(\mathbf{x}),
\label{eq:HCL}
\end{equation}
where the corresponding term is omitted if \(\mathcal{R}\) or \(\mathcal{A}\) is empty in a mini-batch.

\subsection{Model Training and Inference}
\label{sec: Model training and inference}

Combining the supervised classification loss (Eq.~\eqref{eq:CLS}), the
domain-adversarial alignment loss (Eq.~\eqref{eq:DOM}), the domain-aware
coherence regularization (Eq.~\eqref{eq:DCR}), and the hierarchical
contrastive loss (Eq.~\eqref{eq:HCL}), the overall training objective of
DAHCL is formulated as
\begin{equation}
\label{eq:total}
\mathcal{L} = \mathcal{L}_{\mathrm{sup}} + \lambda_1 \mathcal{L}_{\mathrm{dom}} + \lambda_2 \mathcal{L}_{\mathrm{HCL}} + \lambda_3 \mathcal{L}_{\mathrm{DCR}},
\end{equation}
where $\lambda_1$, $\lambda_2$, and $\lambda_3$ are trade-off hyperparameters that balance the contributions of the respective loss terms.

The complete training and inference procedure of DAHCL is summarized in Algorithm~\ref{alg:dahcl}. During training, the feature extractor $\mathcal{F}$, the classifier $G_C$, the domain discriminator $G_D$, and the domain knowledge distiller $M_{\phi}$ are jointly optimized by minimizing the composite objective in Eq.~\eqref{eq:total}. It is worth noting that the DPAS vectors $\mathbf{a}_m$ are maintained as exponential moving averages (Eq.~\eqref{eq:dpas2}) rather than recomputed from scratch at each iteration, ensuring stable and efficient tracking of domain-level geometric characteristics throughout training. At inference time, the target DPAS vector $\mathbf{a}_t$ is computed online from incoming test mini-batches, and the domain expert with the highest cosine similarity is selected to produce the final prediction, enabling label-free adaptation to unseen domains.

\begin{algorithm}[t]
    \small
    \caption{Training and Inference of DAHCL}
    \label{alg:dahcl}
    \begin{algorithmic}[1]
    \Input Labeled source domain $\mathcal{D}_1^s$, unlabeled source domains $\{\mathcal{D}_m^s\}_{m=2}^M$, loss weights $\lambda_1, \lambda_2, \lambda_3$.

    \Output Trained feature extractor $\mathcal{F}$, classifier $G_C$, domain discriminator $G_D$, domain knowledge distiller $M_{\phi}$, DPAS vectors $\{\mathbf{a}_m\}_{m=2}^M$.

\Statex \textbf{--- Training Stage ---}
\State Initialize all network parameters.
\For{each training iteration}
    \State Sample mini-batches from $\mathcal{D}_1^s$ and $\{\mathcal{D}_m^s\}_{m=2}^M$
    \State Extract features $\mathbf{h} = \mathcal{F}(\mathbf{x})$, embeddings $\mathbf{z} = \phi(\mathbf{h})$
    \State Compute supervised loss $\mathcal{L}_{\mathrm{sup}}$ on $\mathcal{D}_1^s$ via Eq.~\eqref{eq:CLS}
    \State Compute domain-adversarial loss $\mathcal{L}_{\mathrm{dom}}$ via Eq.~\eqref{eq:DOM}
    \Statex \hspace{\algorithmicindent}\textit{\% Domain-Aware Learning (DAL)}
    \State Update DPAS vectors $\mathbf{a}_m$ for each unlabeled domain
    \State Construct domain-aware expert classifiers $\{C_m\}_{m=2}^M$
    \State Compute domain-aware coherence regularization $\mathcal{L}_{\mathrm{DCR}}$ via Eq.~\eqref{eq:DCR}
    \Statex \hspace{\algorithmicindent}\textit{\% Hierarchical Contrastive Learning (HCL)}
    \State Partition unlabeled samples into $\mathcal{R}$, $\mathcal{A}$, $\mathcal{U}$ by dynamic thresholding
    \State Compute hierarchical contrastive loss $\mathcal{L}_{\mathrm{HCL}}$ via Eq.~\eqref{eq:HCL}
    \State Update parameters by minimizing:
    \Statex \quad $\mathcal{L} = \mathcal{L}_{\mathrm{sup}} + \lambda_1 \mathcal{L}_{\mathrm{dom}} + \lambda_2 \mathcal{L}_{\mathrm{HCL}} + \lambda_3 \mathcal{L}_{\mathrm{DCR}}$
\EndFor

\Statex \textbf{--- Inference Stage ---}
\State Compute target-domain DPAS $\mathbf{a}_t$ online from test mini-batches
\State Select best-matching expert: $m^* = \arg\max_m \rho_m$, where $\rho_m = \frac{\mathbf{a}_t^\top \mathbf{a}_m}{\|\mathbf{a}_t\|_2 \|\mathbf{a}_m\|_2}$
\State Predict: $\hat{y} = \arg\max_k [\boldsymbol{\ell}_{m^*}(\mathbf{x})]_k$

\end{algorithmic}
\end{algorithm}

\section{Experimental results and analysis}  \label{sec: experiments}

In this section, we present the experimental evaluation of the proposed DAHCL framework. The source code is available at \url{https://github.com/JYREN-Source/DAHCL}.

\subsection{Experimental setup}
\subsubsection{Datasets and preprocessing}

To verify the effectiveness of the proposed method, we conducted comprehensive experiments on three public datasets collected under different operating conditions.

\textbf{(1) CWRU dataset:} The CWRU rolling bearing dataset, a widely used benchmark for rotating machinery fault diagnosis \cite{smith2015rolling}, is publicly available from Case Western Reserve University\footnote{\url{https://engineering.case.edu/bearingdatacenter}}. The evaluated bearing, SKF 6205-2RS JEM, was tested under four health states: normal, inner-race fault, ball fault, and outer-race fault. Faults with defect diameters of 0.007, 0.014, and 0.021 inches were artificially introduced using electric discharge machining. Experiments were conducted under four operating conditions, i.e., condition 0 (0 HP), condition 1 (1 HP), condition 2 (2 HP), and condition 3 (3 HP), with a sampling frequency of 48 kHz. Vibration signals were acquired from the drive end of the bearing under all conditions. Table \ref{tab:fault_label_CWRU} summarizes the fault categories and the corresponding sample quantities.

\begin{table}[!t]  
\small
\caption{Fault categories and sample quantities for the CWRU dataset under each condition}  
\setlength{\tabcolsep}{0pt}    
\renewcommand{\arraystretch}{1.1}
\begin{tabular*}{\columnwidth}{@{\extracolsep{\fill}}lccc}  
\toprule  
\textbf{Fault type} & \textbf{Damage size (inch)} & \textbf{Label} & \textbf{Sample quantity} \\
\midrule  
Inner fault & 0.007 & I1 & 210 \\
            & 0.014 & I2 & 210 \\
            & 0.021 & I3 & 210 \\[0.5ex]  
Outer fault & 0.007 & O1 & 210 \\
            & 0.014 & O2 & 210 \\
            & 0.021 & O3 & 210 \\[0.5ex]  
Ball fault  & 0.007 & B1 & 210 \\
            & 0.014 & B2 & 210 \\
            & 0.021 & B3 & 210 \\[0.5ex]  
Normal      & --    & N  & 210 \\
\bottomrule  
\end{tabular*}  
\label{tab:fault_label_CWRU}  
\end{table}

\textbf{(2) PU dataset:} The Paderborn University (PU) bearing fault dataset \cite{lessmeier2016condition} contains vibration signals sampled at 64 kHz from a test rig composed of a measurement shaft, a rolling bearing test module, an electric motor, a flywheel, and a load motor\footnote{\url{https://mb.uni-paderborn.de/kat/forschung/kat-datacenter/bearing-datacenter}}. Faults are categorized as either artificially induced or naturally developed through accelerated lifetime testing, covering single, repetitive, and compound damage modes. In this study, only naturally degraded bearing data are used. The data were collected under four operating conditions: condition 0 (0.7 Nm, 1000 N, 1500 rpm), condition 1 (0.7 Nm, 1000 N, 900 rpm), condition 2 (0.1 Nm, 1000 N, 1500 rpm), and condition 3 (0.7 Nm, 400 N, 1500 rpm), where the three values denote load torque, radial force, and rotational speed, respectively. The corresponding fault categories are listed in Table \ref{tab:fault_label_PU}.

\begin{table}[!t]  
    \small
    \caption{Fault categories and sample quantities for the PU dataset under each condition}  
\setlength{\tabcolsep}{0pt}   
\renewcommand{\arraystretch}{1.1}
\begin{tabular*}{\columnwidth}{@{\extracolsep{\fill}}ccccc} 
\toprule  
\begin{tabular}[c]{@{}c@{}}\textbf{Damage}\\\textbf{mode}\end{tabular} & \begin{tabular}[c]{@{}c@{}}\textbf{Damage}\\\textbf{position}\end{tabular} & \textbf{Combination} & \begin{tabular}[c]{@{}c@{}}\textbf{Label}\end{tabular} & \begin{tabular}[c]{@{}c@{}}\textbf{Sample}\\\textbf{quantity}\end{tabular} \\
\midrule  
FP  & OR        & S & KA04 & 250 \\
PDI & OR        & S & KA15 & 250 \\
FP  & OR        & R & KA16 & 250 \\
FP  & OR        & S & KA22 & 250 \\
PDI & OR        & R & KA30 & 250 \\
FP  & IR(+OR)   & M & KB23 & 250 \\
FP  & IR(+OR)   & M & KB24 & 250 \\
PDI & OR + IR   & M & KB27 & 250 \\
FP  & IR        & M & KI14 & 250 \\
FP  & IR        & S & KI16 & 250 \\
FP  & IR        & R & KI17 & 250 \\
FP  & IR        & S & KI18 & 250 \\
FP  & IR        & S & KI21 & 250 \\
\bottomrule  
\end{tabular*}  
\label{tab:fault_label_PU}  
\end{table}

\textbf{(3) JUST dataset:} The JUST slewing bearing fault diagnosis dataset \cite{zhou2024study} provides vibration and acoustic emission signals collected from a slewing bearing test rig at Jiangsu University of Science and Technology\footnote{\url{https://data.mendeley.com/datasets/hwg8v5j8t6/1}}. The test specimen is a type-111.10.100 single-row crossed cylindrical roller slewing bearing, and the signals were acquired at 50 kHz using Kistler accelerometers and acoustic emission sensors. The dataset includes one healthy condition and three single-fault modes, collected under four operating conditions: condition 0 (2 rpm, 0 N), condition 1 (2 rpm, 30 N), condition 2 (2 rpm, 60 N), and condition 3 (6 rpm, 0 N), where the two values denote rotational speed and overturning moment, respectively. The fault categories and sample allocations are listed in Table~\ref{tab:fault_label_JUST}.

\begin{table}[!t]
    \small
    \caption{Fault categories and sample quantities for the JUST dataset under each condition}
    \setlength{\tabcolsep}{0pt}
    \renewcommand{\arraystretch}{1.1}
    \begin{tabular*}{\columnwidth}{@{\extracolsep{\fill}}cccc}
    \toprule
    \textbf{Fault type} & \textbf{Damage size} & \textbf{Label} & \textbf{Sample quantity} \\
    \midrule
    Healthy      & --             & N  & 1000 \\
    Inner ring   & Depth: 1 mm    & I  & 1000 \\
    Outer ring   & Depth: 1 mm    & O  & 1000 \\
    Rolling element & 1 roller worn & B1 & 1000 \\
    \bottomrule
    \end{tabular*}
    \label{tab:fault_label_JUST}
\end{table}

For all three datasets, the raw vibration signals are segmented into non-overlapping samples with a segment length of 1024 points. A non-redundant pairwise protocol is then adopted to construct the semi-supervised domain generalization tasks. Specifically, one operating condition is used as the labeled source domain, two conditions are used as unlabeled source domains, and the remaining condition is treated as the unseen target domain for evaluation. The detailed task settings are summarized in Table~\ref{tab:tasks}, where W, T, and J denote the task indices for the CWRU, PU, and JUST datasets, respectively, and C0--C3 denote the operating conditions.

\begin{table}[!t]
    \small
    \caption{Semi-supervised domain generalization fault diagnosis tasks}
    \setlength{\tabcolsep}{0pt}
    \renewcommand{\arraystretch}{1.1}
    \begin{tabular*}{\columnwidth}{@{\extracolsep{\fill}}cccccc}
    \toprule
    \begin{tabular}[c]{@{}c@{}}\textbf{Labeled}\\\textbf{source}\end{tabular} &
    \begin{tabular}[c]{@{}c@{}}\textbf{Unlabeled}\\\textbf{source}\end{tabular} &
    \begin{tabular}[c]{@{}c@{}}\textbf{Unseen}\\\textbf{target}\end{tabular} &
    \begin{tabular}[c]{@{}c@{}}\textbf{CWRU}\\\textbf{task}\end{tabular} &
    \begin{tabular}[c]{@{}c@{}}\textbf{PU}\\\textbf{task}\end{tabular} &
    \begin{tabular}[c]{@{}c@{}}\textbf{JUST}\\\textbf{task}\end{tabular} \\
    \midrule
    C0 & C2, C3 & C1 & W1  & T1  & J1  \\
    C0 & C1, C3 & C2 & W2  & T2  & J2  \\
    C0 & C1, C2 & C3 & W3  & T3  & J3  \\
    C1 & C0, C3 & C2 & W4  & T4  & J4  \\
    C1 & C0, C2 & C3 & W5  & T5  & J5  \\
    C2 & C0, C1 & C3 & W6  & T6  & J6  \\
    \bottomrule
    \end{tabular*}
    \label{tab:tasks}
\end{table}

To evaluate robustness under noisy environments, Gaussian noise is added to each sample during both training and testing for all tasks. Let $s_{raw}(t)$ denote the original vibration signal. The noisy signal captured by the sensor can then be expressed as $s_{noisy}(t) = s_{raw}(t) + noise(t)$, where $\textit{noise}(t)$ denotes Gaussian noise. The noise intensity is controlled by the signal-to-noise ratio (SNR) \cite{zhou2023rotating}:
\begin{equation}  
\textit{SNR}_{dB} = 10 \log_{10} \left( \frac{P_{signal}}{P_{noise}} \right).  
\end{equation}  
where $P_{signal}$ and $P_{noise}$ denote the power of the original signal and the injected noise, respectively.

\subsubsection{Parameter and training configuration}

As shown in Table~\ref{tab:classifier_arch}, we provide the detailed architectures of the classifier and domain-related modules. The feature extractor is implemented using 1D-ConvNeXt \cite{woo2023convnext}. As reported in Table~\ref{tab:hyperparams} and Table~\ref{tab:trainconfig}, the key hyperparameters and training configurations of DAHCL were determined based on preliminary experiments and empirical experience. Unless otherwise specified, all experiments use the same hyperparameter settings and network architecture. All experiments were implemented in PyTorch based on Python 3.9. The hardware platform consisted of a 14th-generation Intel\textsuperscript{\textregistered} Core\texttrademark~i9 processor and an NVIDIA\textsuperscript{\textregistered} GeForce RTX 4060 GPU.
\begin{table}[t]
\centering
\caption{Detailed architecture of model components in DAHCL}
\label{tab:classifier_arch}
\renewcommand{\arraystretch}{1.1}
\begin{tabular}{llll}
\toprule
\textbf{Module} & \textbf{Input} & \textbf{Layers} & \textbf{Output} \\
\midrule
Feature extractor & $(B, 1, 1024)$ & ConvNeXt \cite{woo2023convnext} & $(B, 320, 32)$ \\

\addlinespace[2pt]
Projection layer & $(B, 320, 32)$ & GAP/FC & $(B, 64)$ \\
Classifier head & $(B, 64)$ & ReLU/FC & $(B, K)$ \\

\addlinespace[2pt]
\multirow{2}{*}{Domain discriminator}
  & $(B, 320, 32)$ & GAP/FC & $(B, 64)$ \\
  & $(B, 64)$ & ReLU/FC & $(B, M)$ \\

\addlinespace[2pt]
\multirow{2}{*}{Domain distiller}
  & $(K)$ & FC/ReLU & $(K, 128)$ \\
  & $(K, 128)$ & FC & $(K, 64)$ \\
\bottomrule
\end{tabular}
\end{table}

\begin{table}[t]
\centering
\caption{Key hyperparameter settings used in DAHCL}
\label{tab:hyperparams}
\renewcommand{\arraystretch}{1.1}
\begin{tabular}{lll}
\toprule
\textbf{Parameter} & \textbf{Description} & \textbf{Value} \\
\midrule
$\mu$ & EMA momentum for DPAS & $0.1$\\
$\varepsilon$ & Amplitude in DAL & $0.1$\\
$T$ & Temperature in HCL & $0.07$\\
$\eta_1,\eta_2$ & Quantile ratios for thresholds & $0.25,\ 0.75$\\
$\tau_1,\tau_2$ & Upper bounds for thresholds & $0.6,\ 0.9$\\
$\lambda_1$ & Weight of $\mathcal{L}_{dom}$ & $0.2$\\
$\lambda_2$ & Weight of $\mathcal{L}_{HCL}$ & $0.2$\\
$\lambda_3$ & Weight of $\mathcal{L}_{DCR}$ & $0.1$\\
\bottomrule
\end{tabular}
\end{table}

\begin{table}[t]
\centering
\caption{Training configuration for DAHCL}
\label{tab:trainconfig}
\renewcommand{\arraystretch}{1.1}
\begin{tabular}{ll}
\toprule
\textbf{Setting} & \textbf{Values} \\
\midrule
Total training epochs & $2000$\\
Batch size & $32$\\
Optimizer & AdamW\\
Initial learning rate & $1\times 10^{-4}$\\
Minimum learning rate & $1\times 10^{-6}$\\
Learning rate scheduler & CosineAnnealingLR\\
Weight decay & $5\times 10^{-4}$\\
\bottomrule
\end{tabular}
\end{table}

\subsubsection{Comparison methods}
To validate the effectiveness of the proposed method, five representative semi-supervised domain generalization methods are selected for comparison.

\begin{itemize}
    \item Domain-invariant feature fusion network (DIFFN) \cite{ren2023domain2}: DIFFN learns inter-domain-invariant and intra-domain-invariant representations through a dual-branch architecture, and further enhances feature discriminability and generalization via mutual learning and feature divergence maximization.

    \item Contrast-assisted domain-specificity-removal network (CDSRN) \cite{song2024contrast}: CDSRN explicitly separates domain-invariant features from domain-specific ones via a removal branch, and further improves transferable representation learning using a proxy-contrastive enhancement module.

    \item Semi-supervised dynamic generalization network with dual feature enhancement strategy (SDGN) \cite{jiang2025semi}: SDGN introduces a sample-adaptive dynamic feature extractor together with a dual feature enhancement strategy to strengthen transferable feature learning under unseen working conditions.

    \item Domain fuzzy generalization network (DFGN) \cite{ren2023domain}: DFGN jointly exploits domain fuzzy alignment and metric learning to extract domain-invariant yet discriminative representations from partially labeled multi-source domains.

    \item Mutual-assistance semi-supervised domain generalization network (MSDGN) \cite{zhao2023mutual}: MSDGN combines mutual-assistance pseudo-labeling, entropy-based sample purification, and low-rank decomposition to mine domain-invariant features from labeled and unlabeled samples.
\end{itemize}

Accuracy \cite{powers2011evaluation} is adopted as the evaluation metric. All experiments are repeated five times to ensure statistical reliability, and the reported results correspond to the mean accuracy.

\subsection{Comparative Analysis}

Tables~\ref{Comparison_CWRU}, \ref{tab:Comparison_PU}, and \ref{tab:Comparison_JUST} compare DAHCL with five SSDGFD baselines on the CWRU, PU, and JUST datasets under two SNR settings (10 dB and 0 dB). On CWRU, DAHCL achieves 87.65\% at 10 dB, slightly below SDGN (89.07\%), but surpasses it under 0 dB with 82.85\%, indicating stronger robustness to severe noise. At the task level, DAHCL performs particularly well on W1, W2, W3, and W6. Notably, due to the relatively small inter-domain discrepancy in CWRU, all methods show comparable performance, as conventional alignment is largely sufficient in such mild conditions. On PU, where domain discrepancies are much larger, DAHCL achieves 69.92\% at 10 dB, outperforming SDGN (68.31\%) by 1.61\%. At 0 dB, the gap further increases to 3.52\%. DAHCL maintains consistently strong performance across tasks; for instance, it exceeds SDGN by 8.67\% on T3 at 0 dB. Although CDSRN is competitive, it exhibits significant instability. On JUST, the most challenging dataset, all methods degrade noticeably. DAHCL achieves 61.15\% at 10 dB, surpassing SDGN by 1.77\%, with clear advantages on tasks such as J3, J5, and J6 (e.g., 77.54\% on J3). At 0 dB, DAHCL reaches 51.42\%, further improving over SDGN by 2.84\%, and remains consistently competitive on tasks such as J2, J3, and J6. Overall, DAHCL demonstrates consistent superiority across all datasets, particularly under low SNR (0 dB) and large domain gaps (PU and JUST). The training times on CWRU, PU, and JUST are 2632 s, 2895 s, and 3380 s, respectively, indicating moderate computational cost. From 10 dB to 0 dB, DAHCL shows performance drops of 4.80\%, 9.53\%, and 9.73\%, all smaller than those of competing methods, confirming its robustness.

These gains stem from the proposed two-level design. DAL leverages domain geometric characteristics to calibrate pseudo-labels, alleviating bias from shared classifiers under large domain discrepancies. HCL introduces a fuzzy contrastive mechanism to provide soft supervision for uncertain samples, avoiding premature rejection and overly hard assignments. Together, they enable more reliable and effective utilization of unlabeled data, leading to superior cross-domain generalization.

\begin{table*}[!t]
\caption{Comparison of accuracy (\%) across all tasks on the CWRU dataset under two SNR settings}
\label{Comparison_CWRU}
\centering
\footnotesize
\renewcommand{\arraystretch}{1.1}
\setlength{\tabcolsep}{4pt}
\resizebox{\textwidth}{!}{
\begin{tabular}{lccccccc c ccccccc c}
\toprule
\multirow{2}{*}{Method}
& \multicolumn{7}{c}{10 dB}
& \multicolumn{1}{c}{}
& \multicolumn{7}{c}{0 dB}
& \multirow{2}{*}{\shortstack{Training\\time (s)}} \\
\cmidrule(lr){2-8}\cmidrule(lr){10-16}
& W1 & W2 & W3 & W4 & W5 & W6 & Avg.
& &
W1 & W2 & W3 & W4 & W5 & W6 & Avg. & \\
\midrule
DIFFN  & 78.47 & 74.84 & 78.56 & 84.73 & 89.23 & 88.92 & 82.46
       & \vline &
       69.18 & 69.15 & 70.38 & 79.04 & 88.36 & 81.22 & 76.22 & 2320 \\
CDSRN  & 80.73 & 77.38 & 80.41 & 88.27 & \textbf{95.65} & 89.47 & 85.32
       & \vline &
       74.18 & 71.56 & 74.73 & 80.92 & 89.34 & 83.78 & 79.09 & 2868 \\
SDGN   & 83.06 & \textbf{82.62} & 84.78 & \textbf{90.54} & 95.17 & \textbf{95.24} & \textbf{89.07}
       & \vline &
       77.43 & 74.28 & 77.56 & \textbf{85.41} & \textbf{89.68} & 86.92 & 81.88 & 3857 \\
DFGN   & 76.29 & 72.67 & 76.34 & 82.18 & 87.43 & 86.91 & 80.30
       & \vline &
       70.54 & 67.38 & 72.61 & 77.42 & 82.76 & 79.45 & 75.03 & 1902 \\
MSDGN  & 77.35 & 73.82 & 77.49 & 83.36 & 90.74 & 85.63 & 81.40
       & \vline &
       65.12 & 68.14 & 71.38 & 77.96 & 82.92 & 77.74 & 73.87 & 932 \\
DAHCL  & \textbf{83.72} & 79.48 & \textbf{87.29} & 88.38 & 94.87 & 92.18 & 87.65
       & \vline &
       \textbf{78.26} & \textbf{76.72} & \textbf{80.14} & 84.56 & 89.13 & \textbf{88.29} & \textbf{82.85} & 2632 \\
\bottomrule
\end{tabular}
}
\end{table*}

\begin{table*}[!t]
\caption{Comparison of accuracy (\%) across all tasks on the PU dataset under two SNR settings}
\label{tab:Comparison_PU}
\centering
\footnotesize
\renewcommand{\arraystretch}{1.1}
\setlength{\tabcolsep}{4pt}
\resizebox{\textwidth}{!}{
\begin{tabular}{lccccccc c ccccccc c}
\toprule
\multirow{2}{*}{Method}
& \multicolumn{7}{c}{10 dB}
& \multicolumn{1}{c}{}
& \multicolumn{7}{c}{0 dB}
& \multirow{2}{*}{\shortstack{Training\\time (s)}} \\
\cmidrule(lr){2-8}\cmidrule(lr){10-16}
& T1 & T2 & T3 & T4 & T5 & T6 & Avg.
& &
T1 & T2 & T3 & T4 & T5 & T6 & Avg. & \\
\midrule
DIFFN  & 65.06 & 87.49 & 54.20 & 69.23 & \textbf{47.05} & 50.58 & 62.27
       & \vline &
       48.27 & 84.36 & 46.53 & 61.18 & \textbf{41.68} & 49.24 & 55.21 & 2562 \\
CDSRN  & 70.21 & \textbf{96.87} & 57.62 & 74.55 & 44.85 & 58.58 & 67.11
       & \vline &
       51.26 & \textbf{90.82} & 52.50 & 63.46 & 35.42 & 50.70 & 57.36 & 3133 \\
SDGN   & 71.50 & 93.03 & 58.71 & 75.83 & 45.16 & 65.62 & 68.31
       & \vline &
       \textbf{61.17} & 85.53 & 48.76 & 63.87 & 29.73 & \textbf{52.14} & 56.87 & 4273 \\
DFGN   & 63.82 & 85.10 & 52.91 & 67.81 & 31.56 & 50.39 & 58.60
       & \vline &
       44.74 & 83.15 & 41.18 & 57.26 & 20.45 & 47.24 & 49.00 & 2192 \\
MSDGN  & 66.80 & 88.56 & 55.10 & 71.10 & 33.23 & 54.42 & 61.54
       & \vline &
       46.54 & 83.75 & 42.74 & 57.77 & 22.86 & 51.50 & 50.86 & 1025 \\
DAHCL  & \textbf{73.11} & 92.34 & \textbf{63.28} & \textbf{81.90} & 41.15 & \textbf{67.76} & \textbf{69.92}
       & \vline &
       59.32 & 88.21 & \textbf{57.43} & \textbf{70.18} & 37.15 & 50.06 & \textbf{60.39} & 2895 \\
\bottomrule
\end{tabular}
}
\end{table*}

\begin{table*}[!t]
\caption{Comparison of accuracy (\%) across all tasks on the JUST dataset under two SNR settings}
\label{tab:Comparison_JUST}
\centering
\footnotesize
\renewcommand{\arraystretch}{1.1}
\setlength{\tabcolsep}{4pt}
\resizebox{\textwidth}{!}{
\begin{tabular}{lccccccc c ccccccc c}
\toprule
\multirow{2}{*}{Method}
& \multicolumn{7}{c}{10 dB}
& \multicolumn{1}{c}{}
& \multicolumn{7}{c}{0 dB}
& \multirow{2}{*}{\shortstack{Training\\time (s)}} \\
\cmidrule(lr){2-8}\cmidrule(lr){10-16}
& J1 & J2 & J3 & J4 & J5 & J6 & Avg.
& &
J1 & J2 & J3 & J4 & J5 & J6 & Avg. & \\
\midrule
DIFFN  & 51.82 & 55.34 & 65.28 & 61.74 & 39.46 & 58.72 & 55.39
       & \vline &
       42.15 & 45.28 & 56.37 & 51.73 & 30.64 & 49.29 & 45.91 & 3055 \\
CDSRN  & 53.73 & 57.84 & 68.43 & 64.35 & 39.52 & 61.46 & 57.55
       & \vline &
       39.28 & 42.84 & 54.38 & 49.82 & 25.91 & 46.47 & 43.12 & 3870 \\
SDGN   & \textbf{55.42} & \textbf{59.46} & 70.57 & \textbf{66.82} & 40.76 & 63.28 & 59.38
       & \vline &
       \textbf{45.67} & 48.15 & 60.23 & \textbf{53.48} & 31.84 & 52.12 & 48.58 & 5046 \\
DFGN   & 49.28 & 53.92 & 62.64 & 58.18 & 36.24 & 53.93 & 52.36
       & \vline &
       37.84 & 41.56 & 51.38 & 46.29 & 25.73 & 42.18 & 40.83 & 2587 \\
MSDGN  & 52.35 & 56.74 & 66.92 & 63.46 & 40.18 & 60.37 & 56.67
       & \vline &
       39.28 & 43.37 & 54.84 & 50.23 & 28.12 & 47.65 & 43.92 & 1168 \\
DAHCL  & 53.36 & 56.82 & \textbf{77.54} & 63.73 & \textbf{45.18} & \textbf{69.28} & \textbf{61.15}
       & \vline &
       44.72 & \textbf{49.23} & \textbf{69.79} & 52.35 & \textbf{34.64} & \textbf{57.84} & \textbf{51.42} & 3380 \\
\bottomrule
\end{tabular}
}
\end{table*}

\subsection{Ablation study and validation of effectiveness}
\subsubsection{Ablation study}

\begin{figure}[htbp]
    \centering
    \includegraphics[width=1\linewidth]{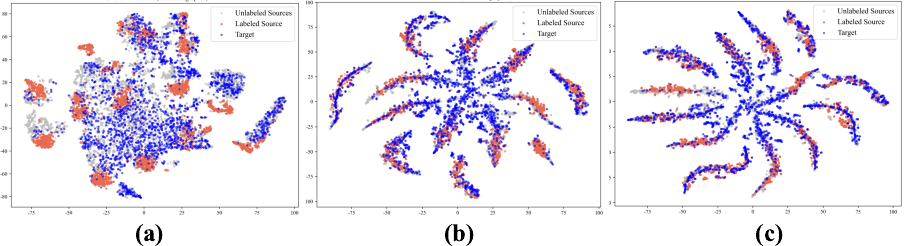}
     \caption{t-SNE visualization of the learned representations on task T4. Subfigures (a)--(c) correspond to M1, M2, and DAHCL, respectively.}
    \label{fig.tsne}
\end{figure}

Table~\ref{tab:ablation_study} presents the ablation results of the key components of the proposed method on the three datasets. Specifically, M1 denotes the backbone model without DAL and HCL; M2 denotes the variant equipped only with HCL while removing DAL; and M3 denotes the model with both DAL and HCL, but without the domain expert selection mechanism during inference. A progressive comparison of these variants clarifies the role of each component. As the baseline, M1 achieves average accuracies of 73.42\%, 42.94\%, and 34.75\% on the CWRU, PU, and JUST datasets, respectively. Its relatively poor performance on PU and JUST suggests that global domain alignment alone is inadequate for scenarios with large inter-domain discrepancies. By introducing HCL into M1, M2 improves the average accuracy by 4.11\%, 9.11\%, and 12.23\% on the three datasets, respectively, confirming the effectiveness of hierarchical contrastive learning. Further comparison between M2 and M3 demonstrates the importance of DAL. Specifically, M3 outperforms M2 by 4.56\%, 7.11\%, and 3.52\% on the three datasets, indicating that DAL can effectively mitigate prediction bias across source domains by incorporating domain geometric characteristics. The impact of removing domain expert selection during inference in M3 is more nuanced. Compared with the full DAHCL model, M3 shows average accuracy drops of 0.76\%, 1.23\%, and 0.92\% on the three datasets, respectively. However, on several tasks (e.g., CWRU-C5 and PU-T2), M3 remains comparable to, or even slightly better than, the full model. This may be because the batch-level dynamic expert selection mechanism is unstable in the early stage of inference adaptation. When the source and target feature distributions are already highly similar, directly using the base classification head can sometimes avoid the adverse effect of inaccurate expert assignment. Fig.~\ref{fig.tsne} presents the t-SNE visualization for task T4. As shown in subfigure (a), M1, which relies only on global domain alignment, fails to learn a stable class-discriminative structure, resulting in blurred inter-class boundaries and evident class mismatch. After incorporating HCL, clearer clustering patterns emerge for both unlabeled source-domain and target-domain samples, as shown in subfigure (b). With both HCL and DAL, DAHCL achieves stronger inter-class separability in subfigure (c), suggesting that DAL further enhances the learning of more accurate class-level domain-invariant representations via domain-geometric constraints, thereby improving cross-domain transfer. Overall, the comparison between DAHCL and its ablation variants demonstrates a clear synergistic effect. Relative to M1, DAHCL improves the average accuracy by 9.43\%, 17.45\%, and 16.67\% on the three datasets, respectively, confirming that the collaboration between DAL and HCL substantially enhances diagnostic robustness under noisy conditions and large domain gaps.

\begin{table*}[!t]
\caption{Ablation study: accuracy (\%) across all tasks on SNR = 0 {dB}.}
\label{tab:ablation_study}
\centering
\footnotesize
\renewcommand{\arraystretch}{1.5}
\setlength{\tabcolsep}{2.5pt}
\begin{tabular}{lccccccc|ccccccc|ccccccc}
\toprule
\multirow{2}{*}{Method} & \multicolumn{7}{c}{CWRU Dataset} & \multicolumn{7}{c}{PU Dataset} & \multicolumn{7}{c}{JUST Dataset} \\
\cline{2-22}
& W1 & W2 & W3 & W4 & W5 & W6 & Avg. & T1 & T2 & T3 & T4 & T5 & T6 & Avg. & J1 & J2 & J3 & J4 & J5 & J6 & Avg. \\
\cline{1-22}
M1 & 68.81 & 67.45 & 70.62 & 75.48 & 83.33 & 76.89 & 73.42 & 33.11 & 78.05 & 36.90 & 46.15 & 22.18 & 41.22 & 42.94 & 28.99 & 31.13 & 50.47 & 35.54 & 22.10 & 40.25 & 34.75 \\
M2 & 72.19 & 71.98 & 74.73 & 79.64 & 84.51 & 83.12 & 77.53 & 45.44 & 85.24 & 44.78 & 60.65 & 30.36 & 45.83 & 52.05 & 38.78 & 42.01 & 66.35 & 47.63 & 33.51 & 52.62 & 46.98 \\
M3 & 78.08 & 75.26 & 79.35 & 83.21 & \textbf{90.02} & 86.63 & 82.09 & 57.99 & \textbf{88.97} & 56.21 & 68.53 & 34.19 & 49.06 & 59.16 & 43.04 & \textbf{47.43} & 69.07 & 50.36 & \textbf{38.40} & 55.72 & 50.50 \\
DAHCL & \textbf{78.26} & \textbf{76.72} & \textbf{80.14} & \textbf{84.56} & 89.13 & \textbf{88.29} & \textbf{82.85} & \textbf{59.32} & 88.21 & \textbf{57.43} & \textbf{70.18} & \textbf{37.15} & \textbf{50.06} & \textbf{60.39} & \textbf{44.72} & 47.23 & \textbf{69.79} & \textbf{52.35} & 36.64 & \textbf{57.84} & \textbf{51.42} \\
\bottomrule
\end{tabular}
\end{table*}

\subsubsection{Effectiveness verification of DAL}

\begin{table}[!t]
\caption{Pseudo-label accuracy (\%) of DAL across unlabeled source domains on the PU dataset ($\mathrm{SNR}=0$ {dB}).}
\label{tab:dal_validation}
\centering
\footnotesize
\renewcommand{\arraystretch}{1.1}
\setlength{\tabcolsep}{3.0pt}
\begin{tabular}{l ccccccc c}
\toprule
Classifier & Domain 
& T1 & T2 & T3 & T4 & T5 & T6 & Avg. \\
\midrule
Base $C$   & $S_1$
& 90.23 & 61.38 & 63.45 & 58.72 & 55.63 & 91.27 & 70.11 \\
DAL $C_1$  & $S_1$
& \textbf{91.19} & \textbf{66.95} & \textbf{68.21} & \textbf{66.84} & \textbf{63.95} & \textbf{92.41} & \textbf{74.93} \\
\midrule
Base $C$   & $S_2$
& 61.54 & 61.67 & 89.75 & 41.18 & 75.41 & 74.58 & 67.36 \\
DAL $C_2$  & $S_2$
& \textbf{66.10} & \textbf{65.79} & \textbf{90.96} & \textbf{49.92} & \textbf{79.37} & \textbf{81.96} & \textbf{72.35} \\
\bottomrule
\end{tabular}
\end{table}

To verify the effectiveness of DAL in mitigating domain bias and improving pseudo-label quality, we report the pseudo-label accuracy on the two unlabeled source domains, \(S_1\) and \(S_2\), for all tasks on the PU dataset. Since the domain-specific expert classifier \(C_m\) outputs a class probability distribution, the pseudo-labels are obtained using the maximum-confidence criterion, i.e., \(\hat{y}=\arg\max_k p_k(x)\). As shown in Table~\ref{tab:dal_validation}, \(C_m\) consistently outperforms the base classifier \(C\) on both unlabeled source domains across all six tasks. Specifically, the average pseudo-label accuracy on \(S_1\) and \(S_2\) increases from 70.11\% and 67.36\% to 74.93\% and 72.35\%, respectively. Averaged over the two unlabeled domains, the pseudo-label accuracy for T1--T6 is improved by 2.77\%, 4.85\%, 2.99\%, 8.43\%, 6.14\%, and 4.26\%, respectively. Notably, DAL tends to yield larger gains in scenarios with more severe domain bias. These results demonstrate that DAL can leverage domain-specific geometric information to calibrate predictions, reduce cross-domain bias, and produce more reliable pseudo-labels.

\subsubsection{Effectiveness verification of HCL}

\begin{table}[!t]
\caption{Effectiveness verification of HCL on the PU dataset under different SNR conditions.}
\label{tab:hcl_validation}
\centering
\footnotesize
\renewcommand{\arraystretch}{1.1}
\setlength{\tabcolsep}{3.5pt}
\begin{tabular}{l cccc cccc}
\toprule
\multirow{2}{*}{Method}
& \multicolumn{4}{c}{SNR = 10 dB}
& \multicolumn{4}{c}{SNR = 0 dB} \\
\cmidrule(lr){2-5} \cmidrule(lr){6-9}
& Acc. & Util. & Bal. & PL Acc.
& Acc. & Util. & Bal. & PL Acc. \\
\midrule
PCL (0.5)
& 49.68 & \textbf{95.12} & 0.58 & 58.36
& 41.52 & 90.85 & 1.46 & 50.12 \\

PCL (0.6)
& 55.42 & 93.34 & \textbf{0.51} & 63.94
& 46.96 & 86.07 & \textbf{0.63} & 55.67 \\

PCL (0.7)
& 60.73 & 90.96 & 0.78 & 71.85
& 49.41 & 84.53 & 0.93 & 57.17 \\

PCL (0.8)
& 64.88 & 89.41 & 0.91 & 75.92
& 53.71 & 82.38 & 1.50 & 65.31 \\

PCL (0.9)
& 58.94 & 85.65 & 1.52 & 72.08
& 41.76 & 72.83 & 2.14 & 58.64 \\

\midrule
HCL (Ours)
& \textbf{69.92} & 94.08 & 0.59 & \textbf{82.41}
& \textbf{60.39} & \textbf{91.74} & 0.66 & \textbf{73.64} \\
\bottomrule
\end{tabular}
\end{table}

\begin{figure}[htbp]
    \centering
    \includegraphics[width=1\linewidth]{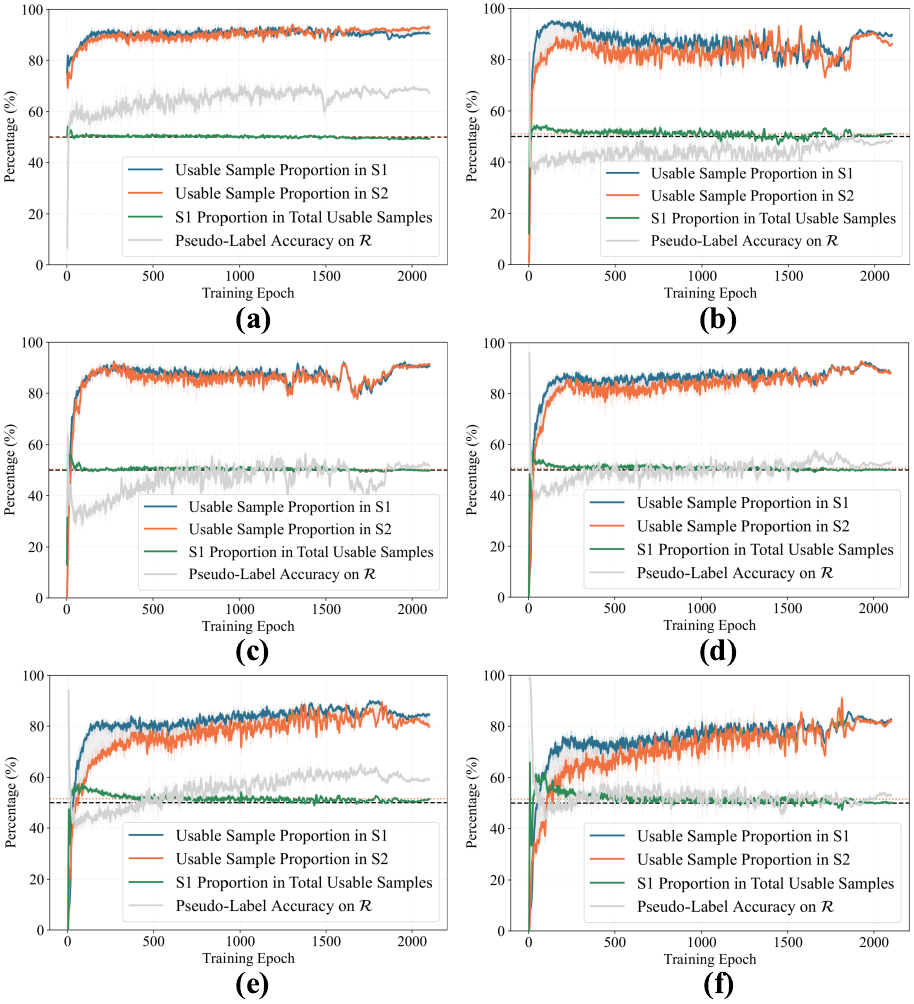}
    \caption{Epoch-wise evolution of four metrics for HCL and PCL on task T4: the proportions of available samples from $S_1$ and $S_2$, the proportion of $S_1$ among all available samples, and the pseudo-label accuracy. Subfigures (a)--(f) correspond to HCL and PCL with $\tau=0.5, 0.6, 0.7, 0.8$, and $0.9$, respectively.}
    \label{fig.epoch}
\end{figure}

\begin{figure*}[htbp]
    \centering
    \includegraphics[width=1\linewidth]{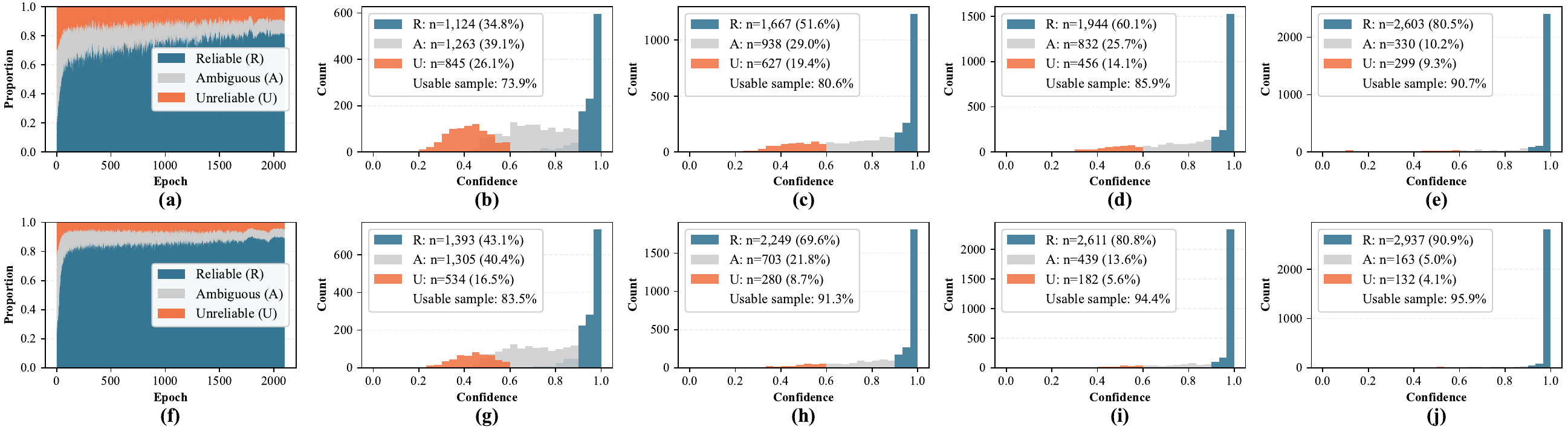}
    \caption{Evolution of domain-specific sample partitions in HCL during training. Subfigures (a) and (f) show the proportions of samples in the $\mathcal{R}$, $\mathcal{A}$, and $\mathcal{U}$ partitions for $S_2$ and $S_1$, respectively, over training epochs. Subfigures (b)--(e) and (g)--(j) present the sampled partition ratios of the two domains at epochs 20, 50, 100, and 2000, respectively.}
    \label{fig.partition_evolution}
\end{figure*}

To verify the effectiveness of HCL under different noise conditions, we compare it with the standard pseudo-label-based contrastive learning method (PCL). For PCL, a sample is assigned a one-hot pseudo-label and included in contrastive learning only when the maximum confidence of its predicted distribution exceeds a threshold \(\tau\); otherwise, the sample is excluded from training. Five fixed threshold settings are considered, i.e., \(\tau \in \{0.5,0.6,0.7,0.8,0.9\}\). The evaluation metrics include test accuracy (Acc.), average sample utilization during training (Util.), inter-domain sample utilization balance (Bal.), and pseudo-label accuracy (PL Acc.). Specifically, Bal. is defined as
\begin{equation}
Bal = \left|100 \times \frac{n_1}{n_1+n_2} - 50\right|,
\end{equation}
where \(n_1\) and \(n_2\) denote the numbers of samples from the two source domains that are actually used for training at the end of training, respectively. A smaller Bal. indicates a more balanced utilization of samples from the two source domains.

Table~\ref{tab:hcl_validation} reports the average results of all methods on the PU dataset under \(\mathrm{SNR}=10\) dB and \(\mathrm{SNR}=0\) dB, while Fig.~\ref{fig.epoch} illustrates their training dynamics on task T4. It can be observed that PCL is sensitive to the threshold choice. A lower threshold increases sample utilization, but introduces more noisy pseudo-labels; in contrast, a higher threshold tends to improve pseudo-label quality, yet reduces the number of training samples and aggravates the imbalance in sample utilization across domains. As the noise level increases, i.e., as \(\mathrm{SNR}\) decreases from 10 dB to 0 dB, the Bal. values of PCL generally increase, indicating that inter-domain imbalance becomes more severe under strong noise, especially at high thresholds. In particular, under \(\mathrm{SNR}=0\) dB, the pseudo-label accuracy of \(\tau=0.9\) is even lower than that of \(\tau=0.8\), while its Bal. reaches the highest value. This suggests that an overly strict filtering strategy causes the available samples in the early stage to be excessively concentrated in \(S_1\), thereby reinforcing the model bias toward \(S_1\) and the labeled source domain, and eventually introducing more erroneous high-confidence pseudo-labels from \(S_2\) in later training. In contrast, HCL achieves the highest classification accuracy and pseudo-label accuracy under both noise conditions while maintaining high sample utilization and stable inter-domain balance, demonstrating a better tradeoff among multiple objectives.

Fig.~\ref{fig.partition_evolution} further presents the partition evolution of the two domains in HCL during training, together with sampled statistics at several key stages. As shown in Fig.~\ref{fig.partition_evolution}(a) and Fig.~\ref{fig.partition_evolution}(f), \(S_2\) and \(S_1\) exhibit markedly different confidence distributions throughout training. In the early stage, more samples from the unlabeled domain that are more similar to the labeled source domain fall into the reliable region, whereas more available samples from the more discrepant domain are concentrated in the ambiguous region. This indicates that the mitigation of inter-domain imbalance by HCL mainly stems from its effective utilization of samples in the ambiguous region. These observations show that the fuzzy contrastive mechanism can exploit most uncertain samples while suppressing noise as much as possible, thereby improving inter-domain balance and alleviating the tradeoff between utilization and accuracy inherent in hard selection strategies.

\subsection{Hyperparameter Sensitivity Analysis}

\begin{figure}[htbp]
    \centering
    \includegraphics[width=1\linewidth]{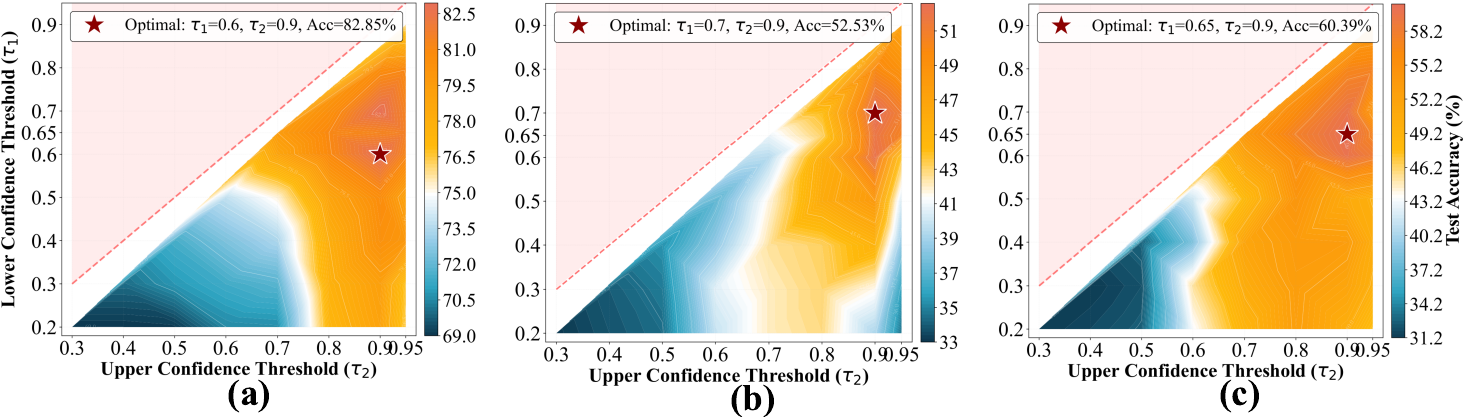}
    \caption{Sensitivity analysis of the confidence thresholds $\tau_1$ and $\tau_2$ on the (a) CWRU, (b) JUST, and (c) PU datasets.}
    \label{fig.sensitivity1}
\end{figure}

\begin{figure}[htbp]
    \centering
    \includegraphics[width=1\linewidth]{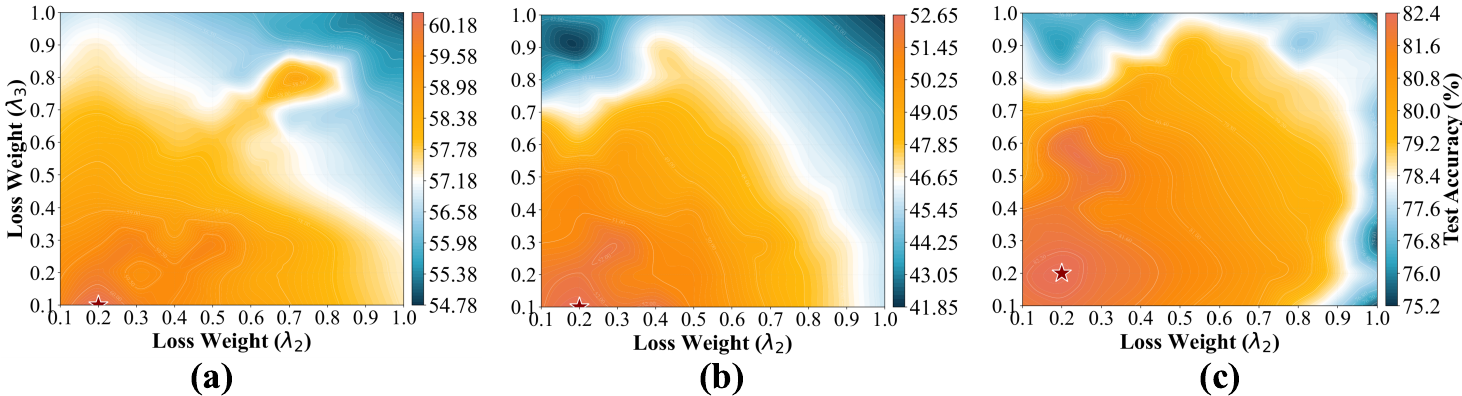}
    \caption{Sensitivity analysis of the loss weights $\lambda_2$ and $\lambda_3$ on the (a) CWRU, (b) JUST, and (c) PU datasets.}
    \label{fig.sensitivity2}
\end{figure}

\begin{figure}[htbp]
    \centering
    \includegraphics[width=1\linewidth]{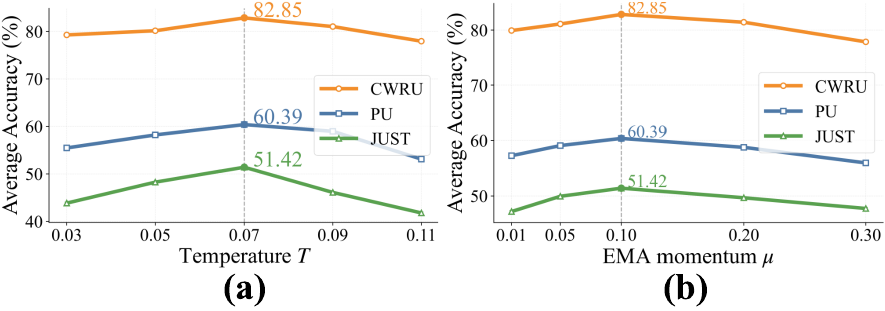}
    \caption{Sensitivity analysis of (a) temperature parameter $T$ and (b) EMA momentum $\mu$, averaged over the three datasets.}
    \label{fig.sensitivity3}
\end{figure}

The proposed method involves several tunable hyperparameters during training. To investigate their influence on diagnostic performance, we analyze the sensitivity of key hyperparameters across the three datasets. The hyperparameters $\tau_1$ and $\tau_2$ serve as confidence thresholds for pseudo-label assignment and thus directly affect the learning of class-discriminative knowledge. Fig.~\ref{fig.sensitivity1} presents the average results across the three datasets. Overall, the ideal range of $\tau_1$ is approximately $[0.58, 0.75]$, whereas that of $\tau_2$ is approximately $[0.85, 0.95]$. In addition, when $\tau_1 < 0.4$ and $\tau_2 < 0.6$, the diagnostic accuracy drops significantly. This is because overly small thresholds tend to introduce incorrect pseudo-labels, which interfere with model learning. By contrast, an excessively large $\tau_1$ often leads to a shortage of available samples, thereby weakening feature learning. The hyperparameters $\lambda_2$ and $\lambda_3$, which control the weights of $\mathcal{L}_{\mathrm{HCL}}$ and $\mathcal{L}_{\mathrm{DCR}}$, are further analyzed. Fig.~\ref{fig.sensitivity2} visualizes the average results. It can be observed that the model is relatively insensitive to $\lambda_2$ and $\lambda_3$. Their optimal values lie approximately in the range of $[0.1, 0.3]$, while performance remains acceptable when they vary within $[0.1, 0.6]$. However, when the weights become excessively large, e.g., within $[0.9, 1.0]$, the diagnostic accuracy decreases markedly. Furthermore, we analyze the influence of the temperature parameter $T$ and the exponential moving average (EMA) momentum parameter $\mu$ used in DPAS. Fig.~\ref{fig.sensitivity3} shows the average results across the three datasets, where Fig.~\ref{fig.sensitivity3}(a) and Fig.~\ref{fig.sensitivity3}(b) correspond to the sensitivity analysis of $T$ and $\mu$, respectively. Overall, the optimal value of $T$ is around $0.07$, with a preferable range of approximately $[0.05, 0.09]$. When $T$ is too small, the interference of noisy samples is easily amplified; when $T$ is too large, the contrastive constraint is weakened, resulting in reduced feature discriminability. The optimal value of $\mu$ is around $0.1$, with a preferable range of approximately $[0.05, 0.20]$. If $\mu$ is too small, the domain prototype update becomes sensitive to mini-batch fluctuations; if $\mu$ is too large, the update becomes overly slow and fails to effectively capture the dynamic feature distribution.

\section{Conclusion} \label{sec:conclusion}

In this paper, we investigated semi-supervised domain generalization fault diagnosis under noisy environments and proposed a unified framework, termed \textit{domain-aware hierarchical contrastive learning}. Unlike existing SSDGFD methods that mainly rely on labeled-source knowledge for pseudo-label generation and handle unlabeled samples in a hard accept-or-discard manner, DAHCL reconsiders the role of unlabeled source domains from two complementary perspectives, namely domain-specific geometry and uncertainty-aware supervision. Specifically, DAHCL leverages domain-specific geometric characteristics to mitigate cross-domain pseudo-label bias, while enabling uncertain samples to contribute to representation learning without imposing unreliable hard labels. To this end, we developed a \textit{domain-aware learning} module to explicitly model source-domain geometric characteristics and calibrate pseudo-label predictions across heterogeneous source domains. We also developed a \textit{hierarchical contrastive learning} module to improve the utilization of unlabeled samples through dynamic confidence stratification and fuzzy contrastive supervision. Extensive experiments on the CWRU, PU, and JUST datasets under multiple SNR settings demonstrated that DAHCL consistently outperforms advanced SSDGFD methods, particularly under severe noise and large domain shifts. Ablation studies and effectiveness analyses further verified the complementarity of DAL and HCL in improving pseudo-label reliability, supervisory quality, and cross-domain sample utilization. Overall, the results indicate that integrating domain-aware bias correction with uncertainty-aware supervision is a promising direction for robust SSDGFD. Future work will extend DAHCL to more complex fault patterns and multi-sensor scenarios, strengthen its theoretical foundations, and explore its applicability to online diagnosis, continual learning, and extremely low-label settings.

\bibliographystyle{IEEEtran}
\bibliography{main}

\end{document}